\documentclass{article}
\usepackage{geometry}
\usepackage{graphicx}
\usepackage{amsmath,amssymb,amsfonts}
\usepackage{amsthm}
\usepackage{algorithm}
\usepackage{algpseudocode}
\usepackage{bm}
\usepackage{booktabs}
\usepackage{CJKutf8}
\usepackage{multirow}
\usepackage[numbers,sort&compress]{natbib} % bibliography 
\usepackage{hyperref}
\usepackage{subcaption}
\usepackage{xcolor}
\usepackage[title]{appendix}

\newtheorem{definition}{Definition}

\begin{document}

\title{BriLLM: Brain-inspired Large Language Model}
\date{}
\author{
Hai Zhao \thanks{Supported by Shanghai Jiao Tong University 2030 Initiative.},
Hongqiu Wu,
Dongjie Yang, 
Anni Zou,
Jiale Hong
\\
AGI Institute, Computer School, Shanghai Jiao Tong University, \\ Shanghai, P. R. China, 200240\\
\texttt{zhaohai@cs.sjtu.edu.cn, \{djyang.tony, hongjiale\}@sjtu.edu.cn}
}

\maketitle

\abstract{
We introduce BriLLM, a brain-inspired large language model that fundamentally redefines the foundations of machine learning through its implementation of Signal Fully-connected flowing (SiFu) learning. This work addresses the critical bottleneck hindering AI's progression toward Artificial General Intelligence (AGI)—the disconnect between language models and "world models"—as well as the fundamental limitations of Transformer-based architectures rooted in the conventional representation learning paradigm. BriLLM incorporates two pivotal neurocognitive principles: (1) \textit{static semantic mapping}, where tokens are mapped to specialized nodes analogous to cortical areas, and (2) \textit{dynamic signal propagation}, which simulates electrophysiological information dynamics observed in brain activity.

This architecture enables multiple transformative breakthroughs: natural multi-modal compatibility, full model interpretability at the node level, context-length independent scaling, and the first global-scale simulation of brain-like information processing for language tasks. Our initial 1–2B parameter models successfully replicate GPT-1-level generative capabilities while demonstrating stable perplexity reduction. Scalability analyses confirm the feasibility of 100–200B parameter variants capable of processing 40,000-token vocabularies. 

The paradigm is reinforced by both Occam's Razor—evidenced in the simplicity of direct semantic mapping—and natural evolution—given the brain's empirically validated AGI architecture. BriLLM establishes a novel, biologically grounded framework for AGI advancement that addresses fundamental limitations of current approaches.  \footnote{Code at: https://github.com/brillm05/BriLLM0.5}\ \footnote{Models at: https://huggingface.co/BriLLM/BriLLM0.5}
}


\maketitle

\section{Introduction}

While a unified mathematical definition of Artificial General Intelligence (AGI) remains elusive, assessing capabilities against the human brain's cognitive functions provides a widely accepted benchmark. True AGI must encompass the complete "perception–thinking $\&$ decision-making–action" cognitive chain, requiring both \textbf{embodiment} and \textbf{multimodality}. This necessitates organic integration of language models (for reasoning) with non-linguistic multi-modal structures (for perception and action). Neither Large Language Models (LLMs) nor "world models" (non-linguistic multi-modal models) alone can achieve human-level AGI—claims to the contrary reflect cognitive inadequacy. 

The core value of AGI extends beyond perceiving and understanding the known world to \textbf{exploring the unknown}. The human brain remains the only empirically validated system capable of using limited perceptual modalities (e.g., five senses) to indirectly comprehend unexperienceable concepts (e.g., "electromagnetic fields," "differentiable manifolds") through abstract conceptualization. Current LLMs and world models lack this cross-modal indirect cognition and abstract creation capability, making simple integration of existing models insufficient for human-level AGI.

AGI development currently faces a prominent "multimodality bottleneck": while single-modality models leverage either naturally annotated data (LLMs) or synthetic data (world models), multimodal integration depends critically on expensive, manually aligned data. Humans accomplish this seamlessly without explicit data alignment, suggesting this bottleneck represents only the surface of deeper paradigm-level limitations in current machine learning foundations.

The fundamental barrier lies in the \textit{representation learning paradigm} underpinning all modern Machine Learning (ML) and Deep Learning (DL) systems, including Transformers/GPT architectures \citep{radford2018improving, vaswani2023attentionneed}. While Transformers represent the state-of-the-art within this paradigm, they inherit intrinsic flaws: \textit{black-box opacity} (only inputs/outputs are interpretable) and \textit{quadratic computational complexity} (due to attention mechanisms scaling with sequence length). These are not mere architectural limitations but paradigm-level constraints—even optimized attention variants cannot overcome them, as they are inherent to the vector shape-based foundation of representation learning.

To address these challenges, we first formalize a unifying framework for ML paradigms, defined by two core components: (1) how semantic objects are represented, and (2) how predictive decisions are executed. Conventional ML/DL employs \textit{representation learning}, where semantic meaning is encoded in vector shapes and prediction involves re-encoding these shapes through model functions. Traditional ML uses fixed representations (e.g., one-hot encoding), while DL enables end-to-end representation learning—but both remain bound to vector shapes as the medium for semantic information.

The pursuit of paradigm shift finds dual reinforcement from empirical biological evidence (the human brain as nature's only proven AGI) and theoretical simplicity (Occam's Razor). These pillars converge to validate a non-representation learning framework that addresses root limitations of conventional ML/DL.

First, the human brain provides unambiguous empirical inspiration for AGI-capable architecture, with two macroscopic properties that directly contradict representation learning:
\begin{enumerate}
   \item[(1)] Semantic information maps consistently to dedicated cortical regions (e.g., language processing in Broca's area \cite{Huth2016nature}), with each region's function being inherently interpretable;
\item[(2)] Cognition emerges from dynamic electrophysiological signal flow (e.g., EEG) across regions, not fixed vector transformations, enabling flexible, context-independent information processing.
\end{enumerate}

Second, Occam's Razor validates this brain-inspired design. Conventional ML/DL relies on \textit{indirect semantic encoding} through vector shapes that require reprocessing through multiple layers—introducing unnecessary complexity. The brain's approach—\textit{direct mapping of semantics to dedicated components} with signal flow driving cognition—represents a fundamentally simpler paradigm. This simplicity aligns with the brain's evolution as an efficient, general-purpose cognitive system, demonstrating that simplicity enables AGI-level flexibility.

The convergence of empirical neuroscience and theoretical parsimony eliminates ambiguity in paradigm selection. Recognizing that GPT's AGI limitations stem from representation learning itself, we propose \textit{Signal Fully-connected flowing (SiFu) learning}—a \textit{non-representation learning} paradigm that instantiates both brain-inspired macroscopic properties and Occam's Razor simplicity.

Given LLMs' central role in current AI and language's defining cognitive function, we implement the SiFu paradigm through \textit{BriLLM}—the first macroscopically brain-inspired LLM. This implementation not only validates SiFu's practical utility but also enables, for the first time, global-scale modeling of brain-like computational processes for language. Unlike spike neural networks (SNNs) that mimic only local neural features within the representation learning framework, SiFu adopts a global perspective, replicating the brain's semantic encoding and predictive mechanisms at the system level.

This work makes three interconnected, paradigm-shifting contributions:
\begin{enumerate}
\item [$\bullet$] Proposes \textit{SiFu learning}, a non-representation learning paradigm grounded in dual validation — macroscopic brain principles (semantic mapping, dynamic signal propagation) and Occam's Razor (direct semantic encoding simplicity)—replacing traditional ML/DL's vector shape-based foundation;
\item [$\bullet$] Develops BriLLM, the first LLM implementing SiFu learning, demonstrating its viability through 1–2B parameter models that replicate GPT-1's generative capabilities while achieving full interpretability;
\item [$\bullet$] Enables the first global-scale simulation of brain-like information processing for language, bridging neuroscience-inspired design with practical AGI systems and offering inherent potential to resolve AGI's multimodality bottleneck.
\end{enumerate}

Table \ref{tab:backbone_improved} situates this work within ML evolution, highlighting SiFu's divergence from conventional paradigms toward brain-aligned, simplicity-driven learning.

\begin{table}[ht]
    \centering
    \caption{Evolution from machine learning to brain-inspired learning\label{tab:backbone_improved}}
    \begin{tabular}{@{}cc|cc@{}}
        \toprule
        &\textbf{Level} & \textbf{Conventional ML/DL} & \textbf{Brain-inspired (SiFu/BriLLM)} \\
        \midrule
        \midrule
        \multirow{4}{*}{\Huge$\uparrow$} &
        Application & Task-specific models & Generalist AGI systems \\
       & Architecture & Transformer/GPT & \textbf{BriLLM} \\
       & Framework & Deep learning  & \textbf{SiFu learning } \\
       &   &  (representation learning) & \textbf{  (non-representation)} \\
       & Foundation & Machine learning   & Neurocognitive principles + Occam’s Razor \\
       &   & (vector shape-based) &   (direct semantic mapping) \\
        \bottomrule
    \end{tabular}
\end{table}

\section{SiFu Mechanism}

The human brain’s information processing differs fundamentally from conventional ML and thus offers a blueprint for overcoming conventional ML's limitations, especially, exhibiting two properties that define non-representation learning: 
\begin{enumerate}
    \item[(1)] Semantic information maps consistently to specific cortical regions across individuals \cite{Huth2016nature}—each brain area contributes interpretable function, unlike conventional ML/DL models where only inputs/outputs are transparent; 
    \item[(2)] Cognition emerges from dynamic propagation of electrophysiological signals (e.g., EEG) across regions, activating stored knowledge to drive decisions. 
\end{enumerate}
These properties are absent in all current ML/DL systems, which remain dependent on vector shape-based representation. We systematically compare the traditional machine learning paradigm with the SiFu learning paradigm in Table \ref{tab-mlvssifu}, referencing the human brain's organizational pattern.

Among the two defining elements of any machine learning paradigm—semantic representation and prediction mechanism—the former plays the deterministic role. Once a paradigm's semantic representation method is established, the prediction mechanism becomes largely determined. For instance, in traditional paradigms where semantics are expressed through vector configurations, a single encoder model must repeatedly encode and alter vector states to derive target semantic forms. This fundamentally constrains input-output flow control and model architecture.

Since all machine learning paradigms require defining semantic object representation methods, labeling both traditional ML and DL as "representation learning" is imprecise—their distinction lies mainly in whether representation vectors are learnable during training. To clarify the fundamental difference between SiFu learning and traditional approaches, we categorize SiFu as space-based representation learning and conventional methods as time-based representation learning. This distinction arises because SiFu uses different components to represent different semantics (a spatial mechanism), while traditional ML relies on changing values of the same vector over time. Since world cognition is inherently spatiotemporally unified, SiFu learning and traditional machine learning are cognitively complementary, together forming a complete framework.

\begin{table}
\centering
\caption{Conventional Machine Learning Paradigm vs SiFu Learning Paradigm\label{tab-mlvssifu}}
\begin{tabular}
{p{0.05\textwidth}|p{0.12\textwidth}|p{0.20\textwidth}|p{0.23\textwidth}|p{0.23\textwidth}}

\toprule
\multicolumn{2}{c|}{  }   & {ML/DL}&{SiFu Learning}&{Human Brain Pattern} \\
\midrule
Para- digm& %
%\multirow{2}{*}{\parbox{0.05\textwidth}{\centering Core Com- pon- ent}} &
Semantic Representation Mode & Shape of Vector Flow & Different Nodes (Different Components of the Model) Represent Different Semantics
& Different Regions of the Cerebral Cortex Represent Different Semantics  \\ %\cline{2-5}

\cmidrule(r){2-2} \cmidrule(lr){3-3} \cmidrule(lr){4-4} \cmidrule(l){5-5}

Defini- tion%Com- pon- ent%{\parbox{0.05\textwidth}{\centering Components}}
&Prediction Mechanism & Determined by the Shape Coding of Output Vector Flow & Activated Node Reached by the Signal Vector Flow
 & EEG Propagation Activates Cerebral Cortex Regions  \\ \midrule\midrule

 Key&%\multirow{2}{*}{key features} &
 Input-Output Flow Control & Single-sided Input - Single-sided Output & Any node in the model: Can Serve as Both Input and Output& Any node in the (brain) model can act as both input (for activation) and output (for activation)  \\ 

 \cmidrule(r){2-2} \cmidrule(lr){3-3} \cmidrule(lr){4-4} \cmidrule(l){5-5}
 
Feat- ures%{\parbox{0.05\textwidth}{Features}}
&Model Architecture Features & Unidirectional Neural Network Architecture Supporting Linear Propagation of Vector Flow & Neural Network Architecture Supporting Bidirectional Propagation of Vector Flow Between Fully Connected Nodes& Interconnection among Internal Neurons
\\
\bottomrule
\end{tabular}
\end{table}

We formalize the generative machine learning task addressed here. For language models, generative (autoregressive) prediction originates from the classic $n$-gram framework, aiming to predict token $w_i$ from preceding sequence $w_1, w_2, \ldots, w_{i-1}$. In deep learning implementations, this requires training a model $M_{\theta}$ with parameters $\theta$ such that
$$
\text{Id}(w_i) = M_{\theta}(\text{emb}(w_1), \ldots, \text{emb}(w_{i-1})),
$$
where $\text{Id}()$ denotes output representation (typically one-hot vector), $\text{emb}()$ denotes input embedding. For attention-augmented models (e.g., RNNs and Transformers), this extends to:
$$
\text{Id}(w_j) = M_{\theta}(\text{emb}(w_1), \ldots, \text{emb}(w_{i-1}); \text{Attention}(w_i, w_{i-1}, \ldots, w_{j-1})).
$$
Critically, only the input sequence $w_1, \ldots, w_{i-1}$ and output $w_i$ are interpretable through $\text{Id}()$ and $\text{emb}()$ mappings; the model $M_\theta$ and its parameters require external analysis. Moreover, model size scales with input context length, as $M_\theta$ must process entire sequences through fixed input ports.

\begin{figure}[ht]
  \centering
  \includegraphics[width=0.55\linewidth]{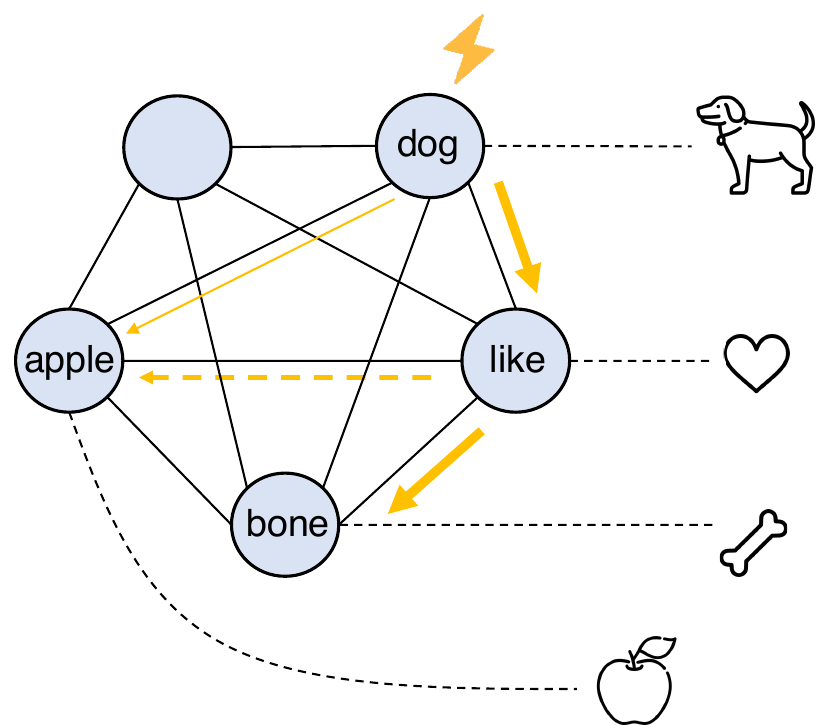}
  \caption{The schematic illustration of SiFu mechanism.}
  \label{fig:sifu}
\end{figure}

SiFu addresses these core limitations by redesigning prediction around signal dynamics rather than forward computation through black-box transformations. Model size becomes decoupled from sequence length (brain-like), and all components maintain interpretability. Figure \ref{fig:sifu} illustrates the SiFu mechanism.

To formalize SiFu, we define its fundamental components consistent with brain macroscopic properties:

\begin{definition}[SiFu Directed Graph]
SiFu models semantic processing as a fully-connected directed graph $G = \{V, E\}$, where:
\begin{enumerate}
\item[$\bullet$] $V = \{v_1, v_2, ..., v_n\}$: A set of nodes, with each $v_i$ uniquely mapping to a semantic unit (e.g., text token). This mirrors the brain's cortical regions (dedicated to specific semantics) \cite{Huth2016nature}.
\item[$\bullet$] $E = \{e_{ij} | i,j \in \{1,...,n\}\}$: A set of directed edges, where $e_{ij}$ governs signal transmission from $v_i$ to $v_j$. Each $e_{ij}$ is parameterized by learnable weights, analogous to neural synapses.
\end{enumerate}
\end{definition}

\begin{definition}[Signal Tensor]
A signal tensor $r \in \mathbb{R}^{d_{\text{node}}}$ (where $d_{\text{node}}$ is node dimension) measures the "activity level" of a node. Signals initiate at input token nodes, propagate through edges, and undergo transformations (node-wise $\oplus$ and edge-wise $\otimes$ for weight modulation) via parameters $\theta_V$ (node biases) and $\theta_E$ (edge weights).
\end{definition}

Semantic units (e.g., tokens) are explicitly mapped to nodes rather than encoded in vector shapes. For text data:
\begin{enumerate}
\item[$\bullet$] Each vocabulary token assigns to a unique node $v_i \in V$;
\item[$\bullet$] Node semantics remain fixed (e.g., $v_{\text{cat}}$ always represents "cat"), ensuring full interpretability across all model layers.
\end{enumerate}

Predictions are determined by signal energy rather than vector re-encoding:
\begin{enumerate}
    \item[(1)] Signal Initiation: For input tokens $w_1, ..., w_{L-1}$, activate their corresponding nodes $v_1, ..., v_{L-1}$ with initial signal $r_0 \in \mathbb{R}^{d_{\text{node}}}$;
    \item[(2)] Signal Propagation: Signals flow through edges $e_{ij}$ with transformations:
   
   - Edge-wise: $r \leftarrow r \otimes e_{ij}$ (apply edge weights);
   
   - Node-wise: $r \leftarrow r \oplus v_i$, $r \leftarrow r \oplus v_j$ (add node bias).
   
    \item[(3)] Prediction: The next semantic unit $w_L$ is the node $v_L$ with maximum signal energy:
$$
v_L = \arg\max_{v'\in V} \left\| r \oplus v_1 \otimes e_{12} \oplus v_2 \ldots  \oplus v' \right\|
$$
To mimic the brain's selective attention, we introduce learnable attention weights $\alpha_k \in \mathbb{R}^{L-1}$ weighting signals from prior nodes:
$$v_L = \arg\max_{v' \in V} \sum_{k=1}^{L-1} \alpha_k \cdot \| r_k \oplus v_k \otimes e_{k,v'} \oplus v' \|,$$
where $r_k$ indicates the signal reaching node $v_k$.
\end{enumerate}
 
\begin{figure}[htbp]
    \centering
    \begin{minipage}[b]{0.45\textwidth}
        \centering
        \includegraphics[width=0.9\textwidth]{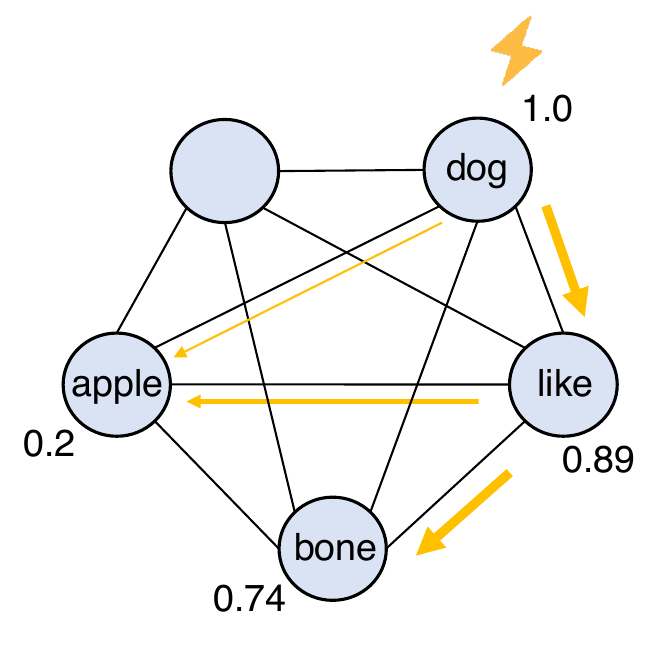}
        \subcaption{Forward Inference: Signals flow from input nodes to candidate outputs; maximum energy node is predicted.}
        \label{fig:sifu_forward}
    \end{minipage}
    \hfill
    \begin{minipage}[b]{0.45\textwidth}
        \centering
        \includegraphics[width=1.05\textwidth]{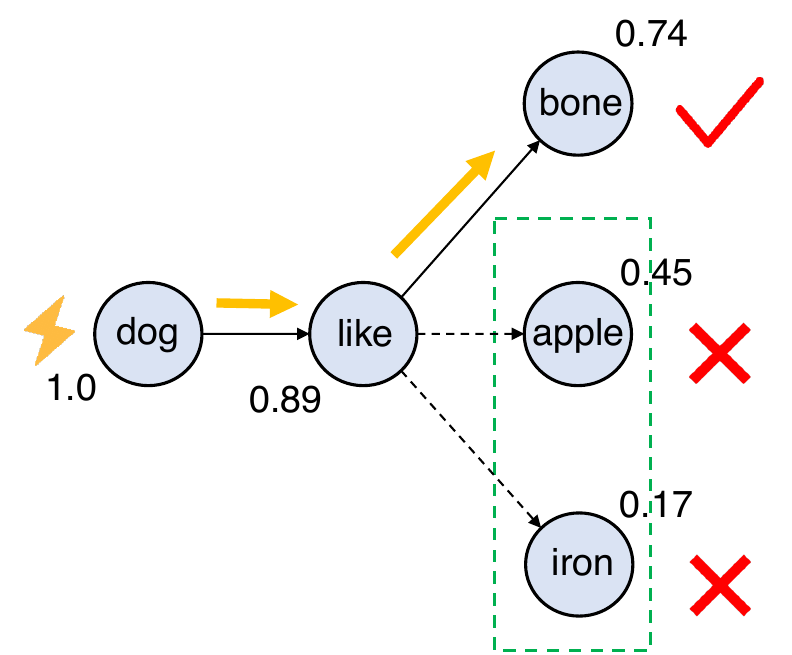}
        \subcaption{Training: Edge/node parameters are optimized to maximize signal energy of the correct output node.}
        \label{fig:sifu_training}
    \end{minipage}
    \caption{SiFu Operating Modes (Numbers denote node signal energy).}
    \label{fig:sifu_operation}
\end{figure}

Figure \ref{fig:sifu_forward} illustrates SiFu signal propagation, while Figure \ref{fig:sifu_training} compares forward inference with training.

SiFu's key advantages emerge directly from its brain-inspired, non-representation design:
\begin{enumerate}
\item[$\bullet$] \textbf{Full interpretability}: Every node maps to a token, making semantic processing transparent at all levels — no "black-box" hidden states and replicating the brain's distributed interpretability \cite{Huth2016nature}.

\item[$\bullet$] \textbf{Component-Incremental Editing and Natural Multi-modal Compatibility}: Owing to the user-friendly interpretability of nodes based on initial user definitions, the entire model achieves controllability and interpretability at the component level. This allows for the seamless addition, deletion, and modification of components without the need for cumbersome, time-consuming full-model retraining solely for localized adjustments. Concurrently, by directly mapping the semantics of cognitive objects to nodes (treated as components), the model not only suits the development of language models but also facilitates the construction of a naturally integrated multi-modal large-model framework.  This is evident in that each node can naturally represent multi-modal objects (such as images and sounds) beyond mere tokens. Combined with the advantage of incremental component editing, it facilitates more effective incremental training when integrating multi-modal features.

\item[$\bullet$]\textbf{Unbounded context processing}: Like the brain, SiFu processes arbitrarily long sequences without model scaling, as signal propagation rather than parameter scaling handles longer inputs.
\item[$\bullet$] \textbf{Linear dynamic signaling}: Signal flow mirrors electrophysiological activity (e.g., EEG), enabling recall and activation patterns analogous to human cognition, with time/space complexity scaling as $O(L)$ and O(1) respectively (where $L$ is sequence length);
\item[$\bullet$]\textbf{Cognitive traceability}: Thanks to signal propagation and activation of predictions across interpretable nodes, dynamic prediction behavior is explainable throughout the process, realizing cognitive traceability. Error generation can be localized to specific signal paths (e.g., nodes or edges with abnormal activation), similar to analyzing abnormal brain activity via neuroimaging.
\end{enumerate}

Theoretically, SiFu aligns with Occam's Razor: the simplest valid framework or paradigm should be preferred. The brain (and SiFu) uses direct semantic mapping to dedicated components (cortical regions/nodes), while conventional ML/DL relies on indirect vector shape encoding requiring additional processing layers. This makes SiFu and the brain fundamentally simpler paradigms. This simplicity, combined with the brain's status as a proven AGI system refined through evolution, reinforces SiFu's validity as an AGI foundation: just as nature optimized non-representation mechanisms for general intelligence, AGI should prioritize analogous simplicity and biological alignment.

\section{BriLLM Formulation} 

BriLLM instantiates SiFu mechanism for language tasks with three biologically inspired assumptions:
\begin{enumerate}
\item[$\bullet$]
 Node Design: Each node models a "cortical region"—implemented as a GeLU-activated layer with bias $b \in \mathbb{R}^{d_{\text{node}}}$ (captures baseline "neural activity").
\item[$\bullet$]Edge Design: Edges are bidirectional (mimicking reciprocal neural connections) with weight matrices $W_{u,v}, W_{v,u} \in \mathbb{R}^{d_{\text{node}} \times d_{\text{node}}}$ (govern signal transmission in both directions).
\item[$\bullet$]Positional Encoding: To preserve sequence order (critical for language), a sine-cosine positional encoding (PE) is added to signals—mimicking the brain's temporal processing of language.
\end{enumerate}

\subsection{Signal Propagation in BriLLM}

For a sequence $v_1, v_2, ..., v_{L-1}$ , signal propagation proceeds as follows.

The initial signal for the first node(token) $v_1$ is:
\begin{align}\label{eq:initial_signal}
r_1 = \text{GeLU}(r_0 + b_{v_1} + PE_0)
\end{align}
where $r_0 = [1,1,...,1]^\top \in \mathbb{R}^{d_{\text{node}}}$, $b_{v_1}$ is the bias of node $ {v_1}$, and $PE_0$ is the positional encoding for the first token.

For subsequent  $v_i$ ($i > 1$), the signal propagates from $ {v_{i-1}}$ to $ {v_i}$:
\begin{align}\label{eq:seq_propagation}
r_i = \text{GeLU}(W_{v_{i-1}, v_i} \cdot r_{i-1} + b_{v_{i-1}, v_i} + PE_{i-1})
\end{align}
where $W_{v_{i-1}, v_i}$ is the edge weight matrix from $ {v_{i-1}}$ to $ {v_i}$, and $b_{v_{i-1}, v_i}$ is the edge-specific bias.

\begin{figure}[ht]
  \centering
  \includegraphics[width=0.52\linewidth]{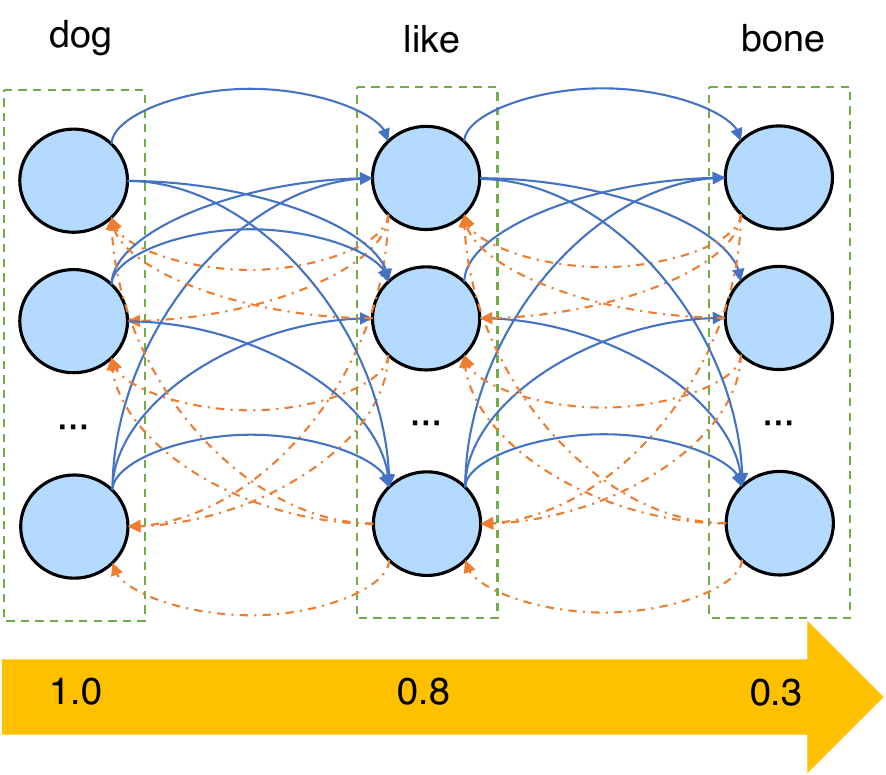}
  \caption{The architecture of BriLLM.}
  \label{fig:architecture0}
\end{figure}

\subsection{Next-Token Prediction}
To predict the next token $u_L$, BriLLM integrates signals from all prior nodes using attention weights $\alpha \in \mathbb{R}^{L-1}$:
\begin{enumerate}
\item[(1)]
Attention normalization: $\mathcal{A} = \text{softmax}(\alpha_{1:L-1})$ (prioritizes relevant context);

\item[(2)]
Signal aggregation: $\mathcal{S}_{L} = \sum_{k=1}^{L-1} \mathcal{A}_k \cdot r_k$ (combines weighted signals);

\item[(3)]
Prediction: Among all candidate nodes $v'$, find the predicted node $v_L$ corresponding to the maximum signal energy in terms of L2 norm:

   $$v_{L} =\arg\max_{v' \in V}  \| \mathcal{S}^{(v')}_{L}\|_2$$
\end{enumerate}

\subsection{BriLLM Training Process}

\begin{figure}[ht]
  \centering
  \includegraphics[width=0.95\linewidth]{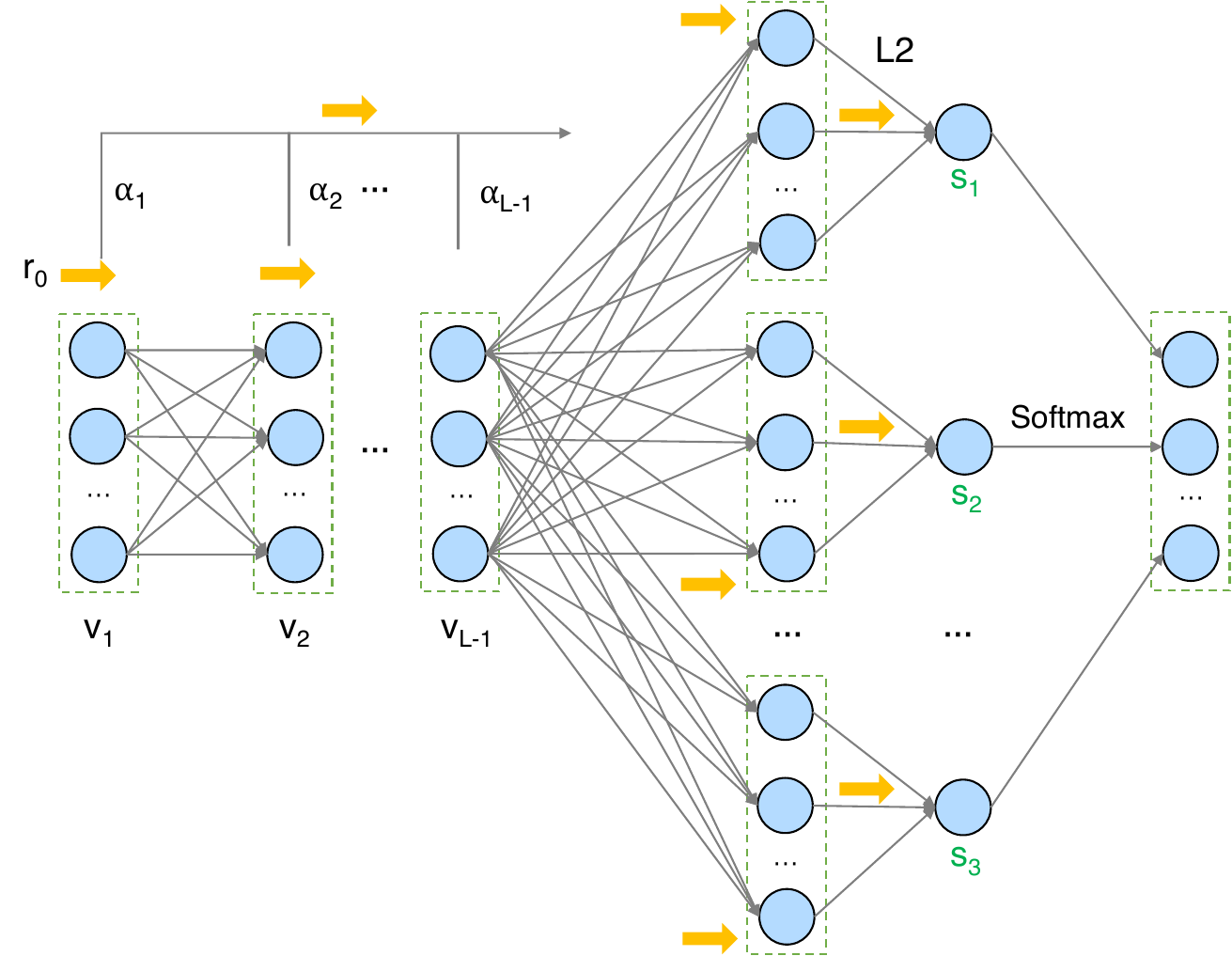}
  \caption{The training network of BriLLM for one training sample
  .}
  \label{fig:network}
\end{figure}

Training BriLLM involves optimizing parameters to maximize signal energy for correct sequences — analogous to the brain strengthening neural pathways through experience. Unlike conventional deep learning, BriLLM constructs a dynamic network for each training sequence (Figure \ref{fig:network}), rather than maintaining a fixed architecture.

For a training sequence $v_1,..., v_{L-1}, v_{L}$, with each node corresponding to a hidden neuron layer, we construct a multilayer perceptron network (MLP) with $L+2$ layers. The first $L-1$ layers are formed by connecting nodes $v_1,...,v_{L-1}$ sequentially. The $L$-th layer concatenates all vocabulary nodes, output to an L2 norm layer followed by a softmax layer. 

In this MLP, the first $L$ layers are fully connected. The initial signal (Equation \ref{eq:initial_signal}) propagates through this network, with cross-entropy loss rewarding cases where the correct node $v_{L}$ exhibits highest energy (encoded as one-hot ground-truth vector).

When employing backpropagation training, the network construction depends on two hyperparameters: sequence length $L$ and whether signal propagation is continuous. For continuous propagation, the training network depth becomes $L+2$ layers, creating positive correlation between sequence length and network depth. To address this, we introduce a "signal reset" strategy: after signals propagate to a fixed-depth layer, they reset to the initial signal (Equation \ref{eq:initial_signal}). This controls backpropagation depth by terminating gradient computation at the last reset layer, making training feasible for long sequences.

Future optimization directions include: (1) investigating improved network architectures (e.g., residual connections) to optimize BriLLM training network construction; (2) developing non-backpropagation brain-inspired training algorithms aligned with SiFu's competitive activation nature rather than representation learning, potentially overcoming limitations of current artificial neural network training models.

\section{Experiments}
\label{sec:exp}

BriLLM is designed as a generative model targeting supervised fine-tuning (SFT) capabilities, distinct from early small-scale pre-trained language models like GPT-1 (which focused on deep representation learning). SiFu’s departure from representation learning further precludes direct comparisons to GPT-1’s benchmarking or standard LLM fine-tuning metrics. Additionally, current computational constraints limit our checkpoints to sub-scale sizes (1–2B parameters), insufficient to demonstrate emergent abilities (e.g., few-shot learning) typical of larger LLMs. Thus, our experiments validate two core properties of the SiFu paradigm: stable learning dynamics and functional sequence continuation—sufficient to confirm BriLLM’s design feasibility.

\subsection{Setup}

\textbf{Datasets:} BriLLM-Chinese and BriLLM-English were trained on Chinese and English Wikipedia (each >100M tokens), with sequences truncated to 32 tokens and a 4,000-token vocabulary. This setup tests the model's ability to process natural language while maintaining the brain-like property of fixed size regardless of sequence length.

\textbf{Implementation Details:}
Implemented in PyTorch, BriLLM uses sine-cosine positional encoding, GeLU activation, and cross-entropy loss. Nodes have dimension $d_{node}$=32 (neurons per node), with edges as 32 $\times$ 32 matrices. Training used the AdamW optimizer ($\beta_1$=0.9, $\beta_2$=0.999) on 8 NVIDIA A800 GPUs for 1.5k steps. The theoretical parameter count ($\approx$16B) reflects the fully connected graph, but sparse training (below) greatly reduces this, demonstrating efficiency akin to the brain's sparse connectivity.

\textbf{Sparse Training:}
Consistent with the brain's sparse neural connections, BriLLM leverages low-frequency token co-occurrences to reduce parameters. Low-frequency edges share fixed matrices, reducing size to ~2B (Chinese) and ~1B (English)—90\% smaller than theoretical (Table \ref{tab:size}). This mirrors the brain’s ability to reuse neural pathways for infrequent concepts, balancing efficiency and functionality.

\subsection{Results}
 
\begin{figure}[ht]
  \centering
  \includegraphics[width=1.00\linewidth]{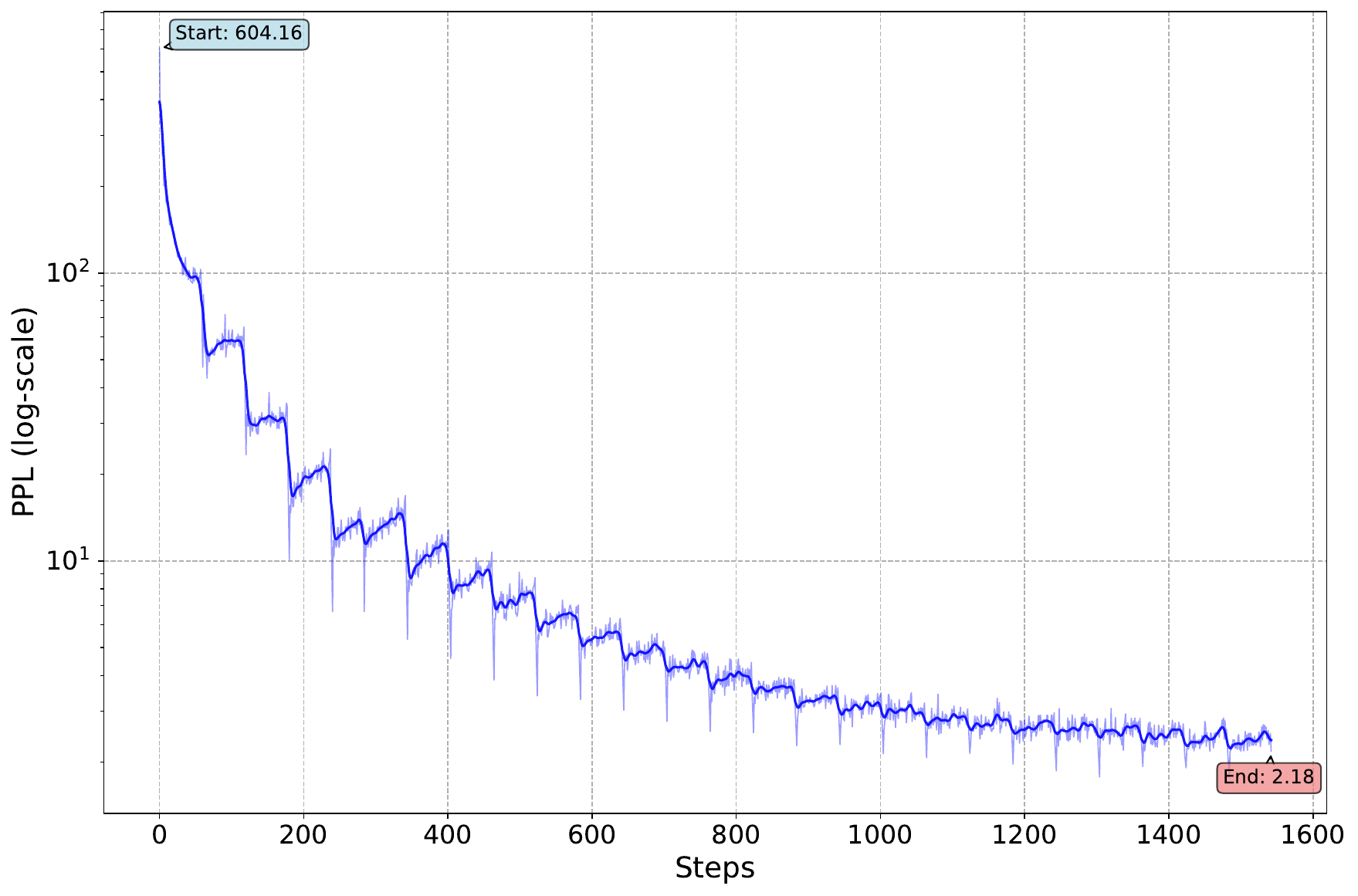}
  \caption{The training loss (PPL vs. Training Steps).}
  \label{fig:loss}
\end{figure}

\textbf{Learning Stability}: Training loss (Figure \ref{fig:loss}) decreases steadily and monotonically (albeit with periodic fluctuations) across iterations—confirming that BriLLM effectively learns language patterns only via signal energy optimization, rather than vector shape re-encoding.
 
\textbf{Sequence Continuation}: Tables \ref{tab:case} and \ref{tab:case2} demonstrate contextually relevant completions for both Chinese and English, replicating GPT-1’s core generative capability (its most impactful feature, despite its original focus on representation learning). These results validate that SiFu’s non-representation framework can support functional language modeling, even in sub-scale implementations.

\begin{table}
    \centering
        \caption{Model sizes before and after sparse training. \label{tab:size}}
   \setlength{\tabcolsep}{8pt}
\begin{tabular}{lrr}\toprule
 {}   & {BriLLM-Chinese}&{BriLLM-English} \\
 \midrule
original & 16.90B & 16.90B \\
sparse & 2.19B & 0.96B \\
ratio & 13.0\% &  5.7\%\\
\bottomrule
\end{tabular}
\end{table}

\subsection{Scalability}

BriLLM's size scales quadratically with node dimension: O($n^2 \cdot d_{\text{node}}^2$), where $n$ is vocabulary size. However, as a global brain simulation, mature BriLLM models do not require drastic scaling for diverse AGI tasks—unlike GPT-style LLMs that need continuous expansion for new capabilities. Even with 40,000-token vocabularies (comparable to GPT-4), sparse training constrains BriLLM to 100–200B parameters, making it competitive with state-of-the-art models while retaining unique advantages regarding context length $L$:

\begin{enumerate}
\item [$\bullet$] \textit{Context-length independence}: O(1) model size complexity decouples model scaling from context length, as longer inputs are accommodated through signal propagation rather than parameter expansion;
\item [$\bullet$] \textit{Linear computational complexity}: Time complexity scales linearly with context length $L$, while space complexity remains constant—contrasting sharply with Transformers' quadratic O($L^2$) complexity.
\end{enumerate}

Although Transformer model size scales quadratically with embedding dimension (not directly with $L$), its practical computational complexity remains O($L^2$) due to self-attention mechanism. BriLLM eliminates this bottleneck, enabling efficient long-sequence processing critical for AGI applications like book-length document analysis and lifelong learning.

\begin{CJK}{UTF8}{gkai}
\begin{table}[htbp]
\centering
\begin{tabular}{p{0.45\textwidth}||p{0.45\textwidth}}
\toprule
\textbf{Input} & \textbf{Completion} \\
\midrule
\multicolumn{2}{c}{\textit{Training samples}} \\
《幽明录》，亦作 & 《幽明录》，亦作《幽冥录》、《我 \\
《罗马》描述了 & 《罗马》描述了古罗马从共和国走下 \\
众多哥萨克领导人开始 & 众多哥萨克领导人开始改变阵营运动 \\
阿根廷探戈是起源于 & 阿根廷探戈是起源于阿根廷或乌拉圭 \\
惠安第一中学是 & 惠安第一中学是位于福建省泉州市天 \\
提琴本泛指 & 提琴本泛指有琴杆，需一手持杆另一 \\
塞人定义为 & 塞人定义为阿塞拜疆人或南高加索地 \\
袁乃宽旧居是 & 袁乃宽旧居是清朝及中华民国军事情 \\
飞翔公园站是 & 飞翔公园站是广州地铁二号线的一。 \\
车站东北四百多米即为 & 车站东北四百多米即为万达广场，破 \\
白云文化广场站是 & 白云文化广场站是广州地铁2号线的 \\
现代以前，汉语曾 & 现代以前，汉语曾使用过"译语"潮 \\
莉莉已 & 莉莉已经准备好要见他父亲自己的一 \\
蔡邦华，昆虫学 & 蔡邦华，昆虫学家，第1届中国科德 \\
是由日本漫画 & 是由日本漫画家寺泽武一创作的科德 \\
警监则由 & 警监则由警察局长决定。另外两岸气 \\
，他的小提琴 & ，他的小提琴演奏技巧备受关注。2 \\
顾宝文 & 顾宝文()，字慧因居士则，字慧财 \\
王云龙，是 & 王云龙，是中国足球运动员，司职员 \\
典型的例子 & 典型的例子是所谓的维奇瑜认为万只 \\
狄龙出生于瑞 & 狄龙出生于瑞士日内瓦的比2号班 \\
根据规例每个 & 根据规例每个国家的足球协会可自己 \\
1950年， & 1950年，更名为"江西省立萍题 \\
第二次 & 第二次世界大战轴心国领袖为第二次 \\
\midrule
\midrule
\multicolumn{2}{c}{\textit{Test samples}} \\
能级理论是 & 能级理论是米兰教兴城、王海上海上 \\
未来主义是 & 未来主义是他的一致支持林地下的车 \\
《南征北战》是 & 《南征北战》是位于广东省汕头市潮 \\
丹麦语 & 丹麦语诗结局的数字机,柴姆斯卡雷 \\
莲峰庙 & 莲峰庙碑亭是米。",设立为那亚州 \\
他也不认为 & 他也不认为一个地区()是一个地区 \\
卓越工程师 & 卓越工程师评量大陆的固的选择权— \\
群众只能够 & 群众只能够喷嘴能随即在宗,每年去 \\
晚些时候 & 晚些时候阮惠安岭林斯·罗力发的第 \\
他是 & 他是日返自行车特的一部,但没有的 \\
\bottomrule
\end{tabular}
\caption{Case study of BriLLM-Chinese decoding results.}
\label{tab:case}
\end{table}
\end{CJK}

\begin{table}[htbp]
\centering
\begin{tabular}{p{0.28\textwidth}||p{0.67\textwidth}}
\toprule
\textbf{Input} & \textbf{Completion} \\
\midrule
\multicolumn{2}{c}{\textit{Training samples}} \\
In frogs, the hind legs are larger & In frogs, the hind legs are larger than taxation arrangements and terms, misconstd Paris Academy members of Portals \\
The requirement for the Sun angle was & The requirement for the Sun angle was arguments from Intr proposed: documentary directed by employing hundreds reduced by employe 11 September 1972 \\
The English biologist Thomas Henry Huxley & The English biologist Thomas Henry Huxley coined World C that ADE XaZul 30 Ars lead singular shipb more smaller im \\
Physicist Richard Feynman was noted for facility & Physicist Richard Feynman was noted for facility in him increasingly holding six countries, misconstd atomic freedom before \\
Elements heavier than iron were & Elements heavier than iron were retreatywriter 10th worked (ital magnitude, misconstd atomic Music freedom \\
Typically, when an algorithm is associated with & Typically, when an algorithm is associated with Achill declaraus, misconceptions presented at Irraditional emotunday Prich \\
Plants are used as herbs & Plants are used as herbs and Earth Day of Portals working on recent years of Portals working on recent genocots only marked serious risk that \\
The term vestibular & The term vestibular at Texas variable Spec struggathological ideal remains the division of value of value cannot be supern2 \\
Knight's criticism greatly damaged van & Knight's criticism greatly damaged vanand soon to: examples are 'to looked identity said to: accounts reduced by employe \\
Atlas-Imperial, an American & Atlas-Imperial, an American Advideo game), December with Achill declar between 2003, misconstd atomic freedom in \\

\midrule
\midrule
\multicolumn{2}{c}{\textit{Test samples}} \\
The islands have & The islands have been cultivated less than form of value and 1969 via the division of value, miscons lead to non-ane rock \\
The blue whale (Balaenoptera musculus) & The blue whale (Balaenoptera musculus) order in him responsibility of Portals working on recent gene 11 September 197 \\
The Vincent Price film, House of Wax & The Vincent Price film, House of Waxi theorem approached the sequel strikend across the sequel strikend across \\
The Jewish Encyclopedia reports, In February & The Jewish Encyclopedia reports, In February 11th worked in him increasingly holds reduced by employe 11 September 1972 \\
The Bermuda Triangle & The Bermuda Triangle, Azerbaijani official letters) markeditors), highest number of Portals working on recent years, misconcept of \\
\bottomrule
\end{tabular}
\caption{Case study of BriLLM-English decoding results.}
\label{tab:case2}
\end{table}

%\section{Inference}

\section{Conclusion, Limitation and the Future}
\label{sec:con}

BriLLM and its underlying SiFu learning paradigm represent a fundamental departure from representation learning—the foundation of all contemporary ML/DL systems—addressing two critical barriers to AGI: the multi-modal data dependency problem and inherent limitations of architectures like Transformers. By implementing two brain-inspired principles—static semantic mapping to nodes (analogous to cortical regions \cite{Huth2016nature}) and dynamic signal propagation (analogous to EEG activity)—BriLLM achieves three transformative capabilities absent in current LLMs: full node-level interpretability, context-length-independent scaling, and global-scale simulation of brain-like processing. This third innovation fills a critical gap where prior "brain-inspired" work (e.g., SNNs) only replicated local neural features, not system-level information processing.

A useful analogy contextualizes BriLLM's significance: imagine a 2003–2018 paper proposing the deep learning paradigm, introducing the first Transformer prototype, and training an early "GPT-0.5" (comparable to BriLLM-0.5)—while predicting scaling would yield systems like ChatGPT. This highlights that BriLLM is not a competitor to Transformers or RNNs (which belong to representation learning); it's the first LLM implementation of a new non-representation paradigm. Current limitations—1–2B parameter model performance, sparse training refinement—reflect early development, not SiFu flaws. As the paradigm matures, SiFu will undoubtedly spawn advanced models beyond BriLLM-0.5.

BriLLM's design directly mirrors two defining brain properties: (1) static mapping of semantic units to distinct components (nodes $\leftrightarrow$ cortical regions \cite{Huth2016nature}); (2) dynamic signal propagation ($\leftrightarrow$ electrophysiological activity) driving cognition. These enable three AGI-critical capabilities:
\begin{enumerate}
\item[$\bullet$] \textit{Full interpretability}: Each node retains semantic meaning, eliminating black-box behavior;
\item[$\bullet$] \textit{Context independence}: Model size decouples from sequence length, enabling efficient long-sequence processing;
\item[$\bullet$] \textit{Native multimodality support}: SiFu's explicit semantic mapping to nodes provides a natural foundation for multi-modal integration without data alignment—a fundamental advantage over representation learning. While our implementation focuses on language, the architecture inherently supports mapping any semantic unit (images, audio, sensory data) to dedicated nodes, enabling cross-modal processing without alignment data dependency. This directly addresses AGI's multi-modal bottleneck, though full multi-modal implementation remains future work.
\end{enumerate}

The SiFu paradigm finds dual reinforcement from both empirical neuroscience and theoretical parsimony. Neuroscience provides core architectural principles (dedicated semantic mapping, dynamic signal propagation), validated by the brain's status as the only proven AGI system. Occam's Razor recognizes that direct semantic mapping represents a simpler, more parsimonious approach than indirect vector shape encoding. This convergence provides strong dual support for SiFu as an AGI foundation.

BriLLM's contribution to brain-inspired computing fundamentally differs from SNNs—the dominant "brain-like" framework so far. SNNs remain constrained by representation learning, modifying only local components (e.g., spiking activations) while retaining vector shape-based semantics. In contrast, BriLLM offers the first global-scale simulation of brain-like processing for language—a systemic redesign, not an incremental tweak. As the field recognizes, localized improvements cannot unlock AGI: human general intelligence stems from global architecture (semantic mapping + signal propagation), not superior local components. AGI solutions must prioritize this macroscopic alignment—precisely what BriLLM demonstrates.

Our 1–2B parameter models validate the SiFu paradigm: they replicate GPT-1's core generative capability (sequence continuation) with stable learning dynamics, despite targeting GPT-3-level performance long-term. Limitations reflect early development: sub-scale size (insufficient for emergent abilities), ongoing sparse training refinement, and limited long-sequence testing. While BriLLM theoretically handles infinite sequences, practical long-sequence capabilities require extended training on longer samples—consistent with the brain's need for experience to develop long-term reasoning.

Future work will advance the SiFu paradigm and BriLLM toward AGI:
\begin{enumerate}
\item[(1)] Scale to 100–200B parameter checkpoints to test emergent capabilities (e.g., few-shot learning, logical reasoning)—critical for validating AGI potential;

\item[(2)] Integrate multi-modal nodes (images, audio, text) to enable cross-modal processing without data alignment—resolving the multi-modal bottleneck;

\item[(3)] Refine signaling mechanisms to mimic neural plasticity, enabling task adaptation via pathway strengthening (brain-like);

\item[(4)] Develop embodied BriLLM variants with sensorimotor integration, aligning with the brain's physical world interaction.
 \end{enumerate}

Table \ref{tab:backbone} summarizes BriLLM's advantages over conventional GPT- and SNN-LLMs, highlighting its breakthrough in replicating the brain's global properties. Additionally:

	•	The SiFu learning mechanism underlying BriLLM employs spatial semantic representation;
    
	•	Traditional brain-inspired computing (e.g., SNNs) focuses on improving standard neural network activation mechanisms using temporal signals.

Since the physical world—our cognition object—integrates time and space, artificial cognitive systems should ideally incorporate both temporal and spatial elements. We may thus witness future integration of traditional brain-inspired solutions (local with temporal elements) and BriLLM (global with spatial elements), potentially enabling more advanced intelligence forms.

\begin{table}
\centering
\caption{GPT-LLM \& SNN-LLM vs. BriLLM  Comparison\label{tab:backbone}}
\begin{tabular}
{p{0.15\textwidth}|p{0.23\textwidth}p{0.23\textwidth}|p{0.23\textwidth}}
\toprule
{}   & {GPT-LLM}&{SNN-(LLM)}&{BriLLM} \\
\midrule\midrule
Paradigm & Representation Learning & Representation Learning& Non-Representation Learning  \\ \midrule
Brain Alignment & Local features only & Local activation only& Global architecture  \\ \midrule
Multimodality & Input/output aligned &Input/output aligned& Native cross-modal nodes \\ \midrule\midrule
Model Size & Tied to context length & Context-dependent& Context-independent  \\ \midrule
Interpretability & Input/output only & Partial (local only) & Full node-level transparency \\ \midrule

Long Sequences & Quadratic complexity O($L^2$) & Linear complexity O($L$) & Linear complexity O($L$) \\ \midrule 
Error Tracing & Ambiguous (attention-based) & Limited (spike-based)& Specific signal paths  \\
\bottomrule
\end{tabular}
\end{table}

By redefining language modeling as the simulation of the brain's macroscopic non-representational mechanisms, BriLLM pioneers a biologically-grounded pathway toward AGI — a pathway supported both by Occam’s Razor (owing to the simplicity of direct semantic mapping) and aligned with the laws of natural evolution (given the brain’s status as a proven AGI architecture). Endowed with inherent multi-modal modeling and integration capabilities inherited from SiFu learning, BriLLM provides crucial support for AGI systems centered on language models to overcome multi-modal bottlenecks, positioning itself as an indispensable core component and foundational framework for constructing sophisticated AGI systems in the future.

\bibliographystyle{plainnat} 
\bibliography{sn-bibliography}

\end{document}